%% file: tunesvdd-main.tex
\def\year{2020}\relax
\documentclass[letterpaper]{article} 
\usepackage{aaai20}  
\usepackage{times}  
\usepackage{helvet} 
\usepackage{courier}  
\usepackage[hyphens]{url}  
\usepackage{graphicx} 
\def\UrlFont{\rm}  
\usepackage{graphicx}  
\title{Active Learning of SVDD Hyperparameter Values}

\author{Holger Trittenbach,\textsuperscript{\rm 1} Klemens B\"ohm,\textsuperscript{\rm 1} Ira Assent\textsuperscript{\rm 2}\\
	\textsuperscript{\rm 1} Karlsruhe Institute of Technology\\ 
	\textsuperscript{\rm 2}	Aarhus University\\ 
	holger.trittenbach@kit.edu, klemens.boehm@kit.edu, ira@cs.au.dk 
}

\usepackage{graphicx}

\usepackage{amsmath,amssymb,amsfonts}
\usepackage{bbm}

\usepackage{booktabs}
\usepackage[inline]{enumitem}
\usepackage{blindtext}
\usepackage{csquotes}
\usepackage{siunitx}

\DeclareMathOperator*{\argmax}{arg\,max}

\usepackage{array}
\newcolumntype{L}[1]{>{\raggedright\let\newline\\\arraybackslash\hspace{0pt}}m{#1}}
\newcolumntype{C}[1]{>{\centering\let\newline\\\arraybackslash\hspace{0pt}}m{#1}}
\newcolumntype{R}[1]{>{\raggedleft\let\newline\\\arraybackslash\hspace{0pt}}m{#1}}

\usepackage{amsthm}

\newtheorem{defn}{Definition}

\usepackage{subcaption}
\usepackage{breqn}
\usepackage{tabularx}

\usepackage[ruled, linesnumbered]{algorithm2e}

\usepackage{color}
\usepackage{balance}

\newcommand{\website}{\url{https://www.ipd.kit.edu/mitarbeiter/lama}}

\begin{document}

\maketitle

\newcommand{\OurMethod}{\text{LAMA}\xspace}

\input{sections/abstract}

\input{sections/introduction}
\input{sections/relatedwork}

\input{sections/method}

\input{sections/experiments}
\input{sections/conclusions}

\section*{Acknowledgments}
This work was supported by the German Research Foundation (DFG), Research Training Group GRK 2153: \textit{Energy Status Data -- Informatics Methods for its Collection, Analysis and Exploitation}.

\bibliographystyle{aaai}
\balance
\bibliography{bibliography}

\end{document}

%% file: sections/abstract.tex
\begin{abstract}
	Support Vector Data Description is a popular method for outlier detection.
	However, its usefulness largely depends on selecting good hyperparameter values -- a difficult problem that has received significant attention in literature.
	Existing methods to estimate hyperparameter values are purely heuristic, and the conditions under which they work well are unclear.
	In this article, we propose \OurMethod (Local Active Min-Max Alignment), the first principled approach to estimate SVDD hyperparameter values by active learning.
	The core idea bases on kernel alignment, which we adapt to active learning with small sample sizes.
	In contrast to many existing approaches, \OurMethod provides estimates for both SVDD hyperparameters.
	These estimates are evidence-based, i.e., rely on actual class labels, and come with a quality score.
	This eliminates the need for manual validation, an issue with current heuristics.
	\OurMethod outperforms state-of-the-art competitors in extensive experiments on real-world data.
	In several cases, \OurMethod even yields results close to the empirical upper bound.
\end{abstract}

%% file: sections/introduction.tex
\section{Introduction}\label{sec:intro}

Support Vector Data Description (SVDD)~\cite{Tax2004-ss} is one of the most popular one-class classifiers for outlier detection.
SVDD builds upon a concise and intuitive optimization problem, and has a wide range of successful applications, including fault monitoring~\cite{Yin2018-jn} and network security~\cite{Gornitz2009-sa,Stokes2008-gn}.
Its core idea is to fit a hypersphere to the data that contains all normal observations; unusual observations fall outside the hypersphere.
SVDD requires to set two hyperparameter values: a kernel function to allow for non-linear decision boundaries and the cost trade-off $C$ that regulates the share of observations that fall outside the hypersphere.
With SVDD, the predominant choice is the Gaussian kernel, which is parameterized by $\gamma$.
In this case, an optimal $\gamma$ reflects the actual complexity of the data, and an optimal $C$ excludes the true share of outliers from the hypersphere.
However, finding out the true complexity and outlier share is challenging, which makes choosing good hyperparameter values difficult.
Moreover, SVDD is sensitive to changes hyperparameter values~\cite{Ghafoori2018-ug}.
It easily over- or underfits the data, which in turn can deteriorate classification quality significantly.

SVDD is usually applied in an unsupervised setting, i.e., the selection of hyperparameter values cannot rely on class label information. 
There is a great variety of heuristics for hyperparameter estimation that use data characteristics~\cite{Khazai2011-dr,Ghafoori2018-ug}, synthetic data generation~\cite{Wang2018-xd,Banhalmi2007-oy}, and properties of the fitted SVDD~\cite{Anaissi2018-yt,Kakde2017-gw} to select a good~$\gamma$.
However, these heuristics do not come with any validation measures or formal guarantees, making it difficult to validate if estimated hyperparameters are indeed a good fit.
Moreover, selecting a suitable heuristic is difficult in the first place, since the intuition of different heuristics may be equally plausible.
This leaves the user with the cumbersome and difficult task of validating the choice of the heuristic and the estimated hyperparameter values.

In this article, we strive for a principled method for SVDD hyperparameter estimation. 
To do away with purely heuristic methods, our idea bases on active learning, i.e., asking users to provide class labels for a few observations that provide grounding for the estimation.

\emph{Challenges.} Developing an active learning method for selecting SVDD hyperparameter values is challenging.
On the one hand, labels give way to using supervised methods for selecting kernel parameters, such as kernel alignment~\cite{Cristianini2002-ud}.
However, a reliable and stable alignment calculation requires a sufficient number of labeled observations~\cite{Abbasnejad2012-xh}.
With active learning, there are only very few labels available, in particular during the first iterations.
Next, current kernel alignment assumes that observations from the same class are similar to each other. 
This assumption may not hold with outlier detection, since outliers are rare, do not have a joint distribution, and may be dissimilar to each other.
So kernel alignment is not applicable without further ado; Section~\ref{sec:alignment-on-sample} illustrates this.
A further challenge is that most conventional active learning strategies are not applicable since they rely on an already parameterized classifier~\cite{Trittenbach2018-zp,trittenbach2019validating}, or focus on fine-tuning of an already parameterized classifier~\cite{Ghasemi2011-tv}.
However, an active learning strategy to estimate hyperparameter values should select observations that are informative of the full data distribution.

\emph{Contributions.}
In this article, we propose \OurMethod (Local Active Min-Max Alignment), an active learning method to select both SVDD hyperparameters $\gamma$ and $C$.
To our knowledge, this is the first active learning method to estimate hyperparameters of SVDD.
It is a principled, evidence-based method and yields a quality score based on the actual class labels obtained by active learning.
This is a key advantage over existing heuristics: \OurMethod does not require manual validation, since its estimations base on labeled observations.

For $\gamma$, we address the challenges in two steps.
First, we propose \emph{locally optimal alignment}, an adapted kernel alignment method based on local neighborhoods.
It confines the calculation to regions where class labels are available.
Second, we propose a novel active learning strategy to explore regions of the data space where class labels are likely to contribute to a reliable alignment calculation.
Estimating $\gamma$ is efficient and widely applicable, since it solely relies on the kernel matrix, and not on any specific model, such as SVDD.
For $C$, we propose a scheme to estimate a feasible lower and upper bound, and then use a grid search to estimate its value.

Empirically, \OurMethod outperforms state-of-the-art heuristics QMS~\cite{Ghafoori2018-ug}, DFN~\cite{Xiao2014-cc} and ADS~\cite{Wang2018-xd} in extensive experiments on real world data.
On several data sets, \OurMethod even yields results close to the empirical upper bound.

%% file: sections/relatedwork.tex
\section{Fundamentals}\label{sec:fundamentals}

We first introduce some notation, and then describe SVDD.

\emph{Notation.}
Let $\mathcal{X} = \{x_1, x_2, \dots, x_N\}$ be a data set with $N$ observations and $M$ dimensions.
Each observation is either an inlier or an outlier, encoded by $y_i \in \{+1 (\text{inlier}), -1 (\text{outlier})\}$.
The number of inliers is $N_{\text{inlier}}$, and of outliers $N_{\text{outlier}}$.
The data can be partitioned into the sets of labeled observations $\mathcal{L}_{\text{inlier}}$ and $\mathcal{L}_{\text{outlier}}$, and of unlabeled observations $\mathcal{U}$.

\emph{SVDD} is a minimum enclosing ball optimization problem.
The basic idea is to fit a hypersphere around the training data by choosing its center $a$ and a radius $R$, such that the boundary is a description of the inliers.
In its most basic form, SVDD is a hard-margin classifier, i.e., all observations lie inside the hypersphere and are considered inliers.
By introducing a trade-off into the optimization problem, SVDD becomes a soft-margin classifier, since some observations, the outliers, can be outside the hypersphere if this decreases the radius significantly.
This is realized by a cost parameter $C \in [0,1]$.
Large values of $C$ make excluding observations from the hypersphere more expensive.
The soft-margin optimization problem is
\begin{equation}\label{eq:svdd-primal}
    \begin{aligned}
        & \underset{a,R,\boldsymbol{\xi}}{\text{minimize}} & & R^2 + C \sum_{i=1}^{N} \xi_i \\
        & \text{subject to}	& & \lVert \Phi(x_i) - a \rVert^2 \leq R^2 + \xi_i, \; i = 1, \ldots, N. \\
        & & & \xi_i \geq 0, \; i = 1, \ldots, N.
    \end{aligned}
\end{equation}
The kernel function $\Phi$ maps observations from the data space into the kernel space, and facilitates arbitrarily shaped decision boundaries.
Since the dual of SVDD depends only on inner products, one can use the kernel trick to replace them with a kernel function $k$.
In our article, we focus on the Gaussian kernel $k(x, x') = e^{- \gamma \lVert x - x' \rVert}$
with kernel bandwidth $\gamma$.
It is by far the most popular kernel used with SVDD.
Existing literature on SVDD hyperparameter selection focuses on choosing $\gamma$ as well.
The kernel function can be interpreted as a similarity measure.
Intuitively, when $\gamma=0$, all observations are projected to the same vector, and their inner product is 1.
For $\gamma \rightarrow \infty$, all observations are orthogonal to each other, and their inner product is 0.
Another way to think about $\gamma$ is the complexity of the decision boundary it induces.
For small values of $\gamma$, the decision boundary in the data space is almost a perfect hypersphere.
Thus, it is less flexible to adjust to the data distribution.
With large values for $\gamma$, the decision boundary is very complex, i.e., the classifier is likely to overfit to the data.
Determining the complexity of the decision boundary is challenging, and makes choosing an appropriate value for $\gamma$ difficult.

The symmetric matrix $K^{N \times N}$ denotes the Gram matrix of the pairwise kernel functions.
We use the shorthand $K(i,j)$ to refer to the entries of $K$.

\section{Related Work}\label{sec:related-work}

Both SVDD hyperparameters depend on each other, i.e., a good value for $C$ depends on the choice of the kernel.
$C$ influences the share of observations that are classified as outlier, and a good value depends on the specific application.
Literature has produced several heuristics to select an appropriate $\gamma$, but the choice of $C$ is often left to the user.
In some cases, the heuristics to select a kernel even require users to initialize $C$ -- a requirement unlikely to be met in practice.

The bulk of methods we present in this section focuses on selecting $\gamma$.
There are three types of heuristics.
The first type is \emph{data-based} selection, which solely relies on data characteristics to estimate $\gamma$, often in a closed formula.
The second type is \emph{model-based} selection, which optimizes for criteria based on the trained model, and thus requires solving SVDD, often multiple times.
The third type of selection heuristics generates \emph{synthetic data} in combination with supervised selection schemes to fit a decision boundary.
In rare cases, when labeled training data is available, one can use plain \emph{supervised selection}, e.g., by using cross validation.

\subsubsection{Data-based Selection}

The simplest data-based estimation methods are formulas to directly calculate $\gamma$. 
There are two rules of thumb by Scott~\cite{scott2015multivariate} and by Silverman~\cite{Silverman1986-cn}. 
They use the number of observations and lower-order statistics to calculate $\gamma$ in a closed formula.
Others propose to estimate $\gamma$ by using the distances between the centers of the outlier and inlier class~\cite{Khazai2011-dr}.
Recent approaches use changes in the neighborhood density of training observations to derive closed formulas for $\gamma$ and $C$~\cite{Ghafoori2018-ug,Ghafoori2016-ux}.

A different approach is to define desired properties of the kernel matrix, and optimize for them by modifying the kernel parameter.
Several such objectives have been proposed: to maximize the coefficient of variance of non-diagonal kernel matrix entries~\cite{Evangelista2007-aa}, to ensure that the kernel matrix is different from the identity matrix~\cite{Chaudhuri2017-bd}, and to maximize the difference between distances to the nearest and to the farthest neighbours of training observations~\cite{Xiao2014-cc}.

\subsubsection{Model-based Selection}

Changes in hyperparameter values modify the optimal solution of the SVDD optimization problem, and the properties of this solution.
Model-based selection strategies fit SVDD for several $\gamma$ values, and select the model which has the desired properties.
A common approach is to define desired geometric properties of the decision boundary.
For instance, one can define criteria on the tightness of a decision boundary, e.g., by estimating whether the decision function is a boundary on a convex set~\cite{Xiao2014-cc}.
A good kernel parameter leads to a decision boundary that is neither too tight nor too loose.
Variants of this approach are to first detect edge points of the data sample~\cite{Xiao2015-sg,Anaissi2018-gv}.
Intuitively, interior points should be far from the decision boundary, and edge points close to the decision boundary.
Thus, one can maximize the difference between the maximum distance of an interior point to the decision boundary and the maximum distance of an edge-point to the decision boundary to balance between tight and loose boundaries.

Others have suggested optimization criteria based on the number of support vectors, i.e., the observations that define the decision boundary.
The number of support vectors tends to increase with more complex decision boundaries.
So one can search for the smallest $\gamma$ such that the number of support vectors are the lower bound imposed by $C$~\cite{Gurram2011-fs}.
A variant of this idea is to decrease $\gamma$ until all support vectors are edge points~\cite{Anaissi2018-yt}.
A different approach is to select the kernel parameter by training SVDD on multiple resamples of the data and then select the $\gamma$ that results in the smallest average number of support vectors over all samples~\cite{Banerjee2006-jv}.

One can also derive objectives directly from the dual objective function of SVDD.
For instance, empirical observations suggest that one can set the second derivative of the dual objective with respect to the kernel parameter to zero to obtain a parameter estimate~\cite{Kakde2017-gw,Peredriy2017-qh}.
Also combinations of support vector count and objective function maximization have been proposed as objectives~\cite{Wang2013-jv}.

\subsubsection{Selection with Synthetic Data}

The core idea of parameter tuning by synthetic data generation is to enhance the training data with labeled artificial observations.
One can then apply supervised parameter tuning, such as grid search and cross-validation, to select parameters that fit best the artificially generated data set.
A benefit is that many of the synthetic data generation methods also provide an estimate for $C$.
However, the success of these methods depends on how well the artificial observations are placed, and whether this placement works well for the data at hand is unclear.
A poor placement can yield parameter values that have very poor classification quality, see Section~\ref{sec:experiments}.

The basic variants generate outliers either uniformly~\cite{Tax2001-ss} or from a skewed distribution~\cite{Deng2007-pc} and estimate the false negative rate for the outliers generated.
To generate outliers more reliably in high-dimensional data, there are adaptations that decompose this problem into first detecting the edge points of the sample and then generating the artificial outliers based on them~\cite{Wang2018-xd,Banhalmi2007-oy}.

\subsubsection{Supervised Selection}

If labeled training data is available, one can use supervised hyperparameter tuning \cite{Tax2004-in,Tran2005-ts,Theissler2013-sr}.
However, these methods are not relevant for our article since with active learning, there initially is no labeled training data available.\\

To conclude, there is a plethora of heuristics available to set the hyperparameter values of SVDD.
However, selecting a suitable heuristic is difficult for several reasons.
For one, there is no objective criterion to compare heuristics.
They do not come with any formal guarantees on their result quality, but offer different intuitions on SVDD, and on motivations for particular estimation strategies.
Respective articles generally do not discuss the conditions under which the heuristics work well.
Next, existing experimental evaluations comprise only a few of the heuristics, and in many cases only a very limited body of benchmark data.
A further important downside of many existing heuristics is that they require to set $C$ manually. 
This makes both a competitive comparison and the application in practice difficult.

%% file: sections/method.tex
\section{Method}\label{sec:method}

In this section, we propose an active learning method to learn hyperparameters values of SVDD.
We first present some preliminaries on kernel learning.
Then we focus on cases when only a few labeled observations are available.
We then present a query strategy to identify observations that are most informative for learning the kernel parameter value.
Finally, we propose a strategy to estimate the cost parameter $C$ based on the set of labels acquired by active learning.

\subsection{Kernel Learning Fundamentals}\label{sec:kernel-learning}

Kernel learning methods construct a kernel matrix or a kernel function from labeled training data, or from pairwise constraints.
The idea is to identify a good kernel given the training data, independent of the classifier.
There are multiple approaches to learn a kernel, e.g., by directly learning the Kernel Matrix (Non-Parametric Kernel Learning)~\cite{Zhuang2011-za} or to learn an optimal combination of multiple kernels, which may differ by type or by parameterization~\cite{Gonen2011-zb}.

With SVDD, the Gaussian kernel is the predominant choice for the kernel.
The Gaussian kernel is parametric, which gives way to learning good parameter values by so-called kernel alignment~\cite{Cristianini2002-ud}.
The idea of kernel alignment is to define an ideal kernel matrix
\begin{equation}\label{eq:K-opt}
    K_{\text{opt}} = yy^{\intercal}
\end{equation} 
using class labels $y$.
The entries of $K_{\text{opt}}$ are $+1$ if observations have the same class label, and $-1$ otherwise.
The alignment between an empirical and ideal kernel matrix is 
\begin{equation}\label{eq:alignment}
    A(K_{\gamma}, K_{\text{opt}}) = \frac{{\langle K_{\gamma}, K_{\text{opt}} \rangle}_{F}}{\sqrt{{\langle K_{\gamma}, K_{\gamma} \rangle}_{F} {\langle K_{\text{opt}}, K_{\text{opt}} \rangle}_{F}}}
\end{equation}
where ${\langle \cdot, \cdot \rangle}_{F}$ is the Frobenius inner product.
Kernel alignment has some desirable theoretical properties~\cite{Wang2015-wo}:
it is \emph{computationally efficient}, i.e., the computation only depends on the number of labeled observations $\mathcal{O}(\lvert \mathcal{L} \rvert^2)$; it is \emph{concentrated} around its expected value, i.e., the empirical alignment deviates only slightly from to the true alignment value; it \emph{generalizes} well to a test set.

Kernel alignment is useful for finding a good kernel parameter.
By using the kernel alignment as an objective, one can search for an optimum~\cite{Wang2015-wo}
\begin{equation}\label{eq:gamma-opt}
    \gamma_{\text{opt}} = \argmax_\gamma \; A(K_{\gamma}, K_{\text{opt}}).
\end{equation}
With outlier detection, calculating the alignment is more difficult, since the class distributions are highly imbalanced.
In this case, the sensitivity of the alignment measure may drop~\cite{Wang2015-wo}.
One remedy is to adjust $y$ by the relative class frequency~\cite{Kandola2002-ze}.
Another method to deal with unbalanced classes is to center the kernel matrix~\cite{Cortes2012-qc}.
Preliminary experiments indicate that relative class frequency adjustment does not improve the alignment calculation in our setting.
We therefore rely on kernel matrix centering.

\subsection{Alignment on Small Samples}\label{sec:alignment-on-sample}

One difficulty of kernel alignment is that it generally requires a large set of labeled examples to define the ideal kernel matrix~\cite{Abbasnejad2012-xh}.
However, with active learning, only very few labels are available, in particular during the first few iterations.
A second difficulty is that user labels may be noisy, i.e., the actual label may differ from the user-provided label.
A reason is that labeling is a subjective assessment, and that users may misjudge and provide a wrong label.
In general, this issue may be negligible, in particular when feedback is correct in most of the cases.
However, noisy labels may impact the kernel alignment significantly when the amount of labeled data is small.

In the following, we propose a method that creates a local alignment to mitigate both of these difficulties.
The idea is to include the local neighborhood of labeled observations in the alignment calculation.
Our methods consists of two steps.
In the first step, we re-label observations based on a majority vote of the labels in their local neighborhood.
The reason for this is two-fold.
On the one hand, this step reduces the influence of noisy labels.
On the other hand, this creates pseudo labels for observations in $\mathcal{U}$ and increases the number of observations for the alignment calculation.
In the second step, we define a locally optimal kernel matrix for the alignment.
That is, we limit the comparison between $K_\gamma$ and $K_{\text{opt}}$ to the relevant entries.

\subsubsection{Preliminaries}
We first introduce some useful definitions.
\begin{defn}[Nearest Neighbors]
    $\text{NN}_k(x)$ are the $k$ closest observations of an observation $x$. 
    We set $\text{NN}_1(x) = \{x\}$.
\end{defn}
\begin{defn}[Reverse Nearest Neighbors]
    $\text{RNN}_k(x)$ is the set of observations that have $x$ as one of their k-nearest neighbors.
    \begin{equation}
       \text{RNN}_k(x) = \{l \; | \; x \in \text{NN}_k(l)\}
    \end{equation}
\end{defn}

\begin{defn}[Symmetric Nearest Neighbors]
    $\text{SNN}_k(x)$ is the set of observations that are k-nearest neighbors of x as well as reverse nearest neighbors of x.
    \begin{equation}
        \text{SNN}_k(x) = \{l \in \text{NN}_k(x) \; | \; x \in \text{NN}_k(l) \}
    \end{equation}
\end{defn}
\subsubsection{Relabeling}
We propose to relabel observations based on their local neighborhood to increase the number of labeled observations, and to reduce the influence of noisy labels.
More specifically, when a user labels an observation $x_i$, this label is propagated to the local neighborhood of $x_i$.
We propose an asymmetric propagation scheme.
When $x_i$ is inlier, the label propagates to the k-nearest neighbors of $x_i$.
So the nearest neighbors of an inlier are deemed inliers as well.
When $x_i$ is outlier, the label propagates to the symmetric nearest neighbors of $x_i$.
The rationale behind this propagation scheme is that the nearest neighbor of an outlier may well be an inlier -- this holds with certainty if there is only one outlier in the data space.
But the nearest neighbors of inliers are likely to also be inliers.
So asymmetric propagation mitigates wrong label propagation.

\begin{figure}[t!]
     \centering
     \includegraphics[scale=1.2]{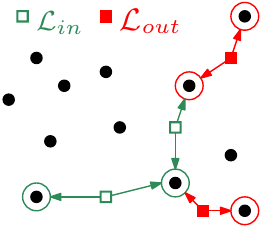}
     \caption{Relabeling with local neighborhoods. The arrows indicate the propagation of class labels to $\text{NN}_{2}$ (green) and $\text{SNN}_2$ (red) neighborhoods of the labeled observations.}
     \label{fig:relabel}
 \end{figure}

After relabeling, one can count how often $x$ occurs as a reverse k-nearest neighbor of labeled inliers, i.e.,
\begin{equation}
    n_{\text{in}}(x) = \sum_{l \in \mathcal{L_{\text{in}}}} \mathbbm{1}_{ \text{NN}_{k}(l) }(x).
\end{equation}
Analogously, the number of times $x$ occurs in symmetric nearest neighbors of outliers is
\begin{equation}
    n_{\text{out}}(x) = \sum_{l \in \mathcal{L_{\text{out}}}} \mathbbm{1}_{ \text{SNN}_{k}(l) }(x).
\end{equation}
Based on these counts, neighborhoods are relabeled based on a majority vote.
The re-labeled pools are
\begin{equation}\label{eq:relabel}
\begin{split}
    & \mathcal{L}_{\text{in}}' = \{x \;|\; \frac{n_{\text{in}}(x)}{n_{\text{in}}(x) \, + \, n_{\text{out}}(x)} > 0.5 \}, \text{and} \\
    & \mathcal{L}_{\text{out}}' = \{x \;|\; 0 < \frac{n_{\text{in}}(x)}{n_{\text{in}}(x) \, + \, n_{\text{out}}(x)} \leq 0.5 \}.
    \end{split}
\end{equation}
The set $\mathcal{U}'$ contains the remaining observations, i.e., the ones that do not occur in neighborhoods of labeled observations.
 Figure~\ref{fig:relabel} illustrates the relabeling.

The optimal kernel matrix based on relabeled observations is
\begin{equation}\label{eq:K-opt-prime}
    K'_{\text{opt}} = y'(y')^\intercal,
\end{equation}
where $y'$ is the label vector after relabeling, cf.\ Equation~\ref{eq:K-opt}.

\begin{figure*}[ht]
     \centering
    \begin{subfigure}[c]{\textwidth}
    \centering
    \includegraphics[scale=0.7]{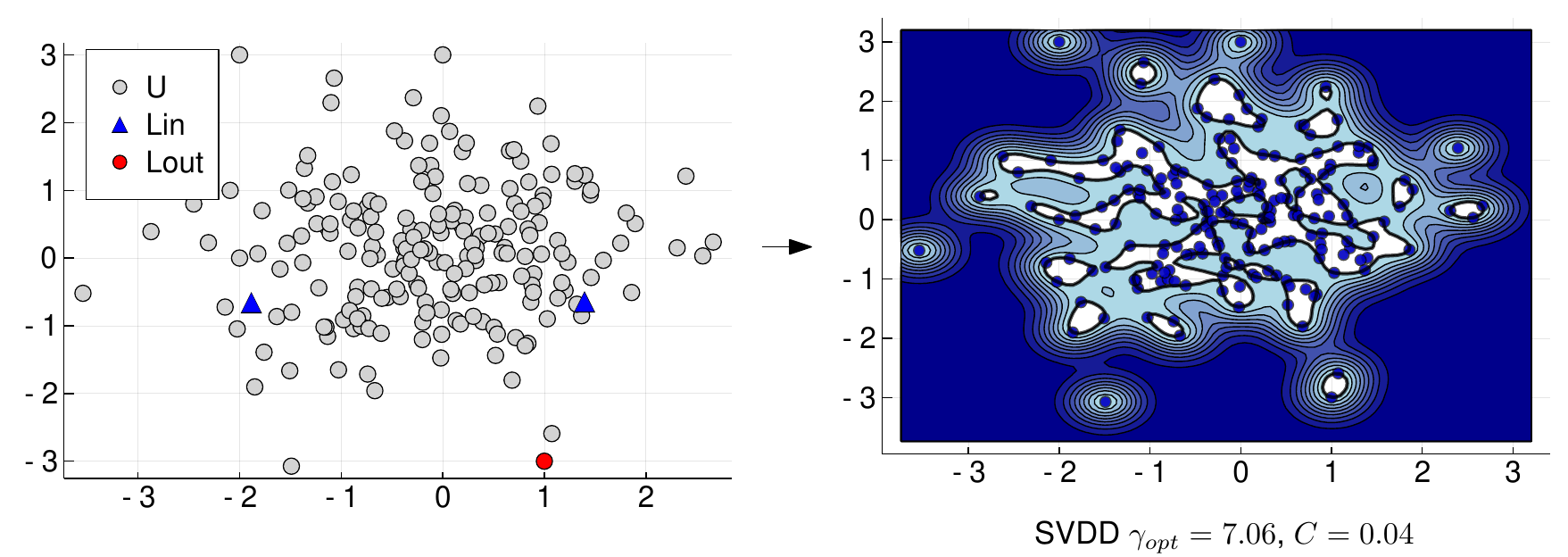}
    \subcaption{Global alignment $\gamma_{\text{opt}}$ based on $L_{\text{in}}$ and $L_{\text{out}}$.}
    \label{fig:stability-illustration-global}
    \end{subfigure}
    \hfill
    \begin{subfigure}[c]{\textwidth}
        \centering
        \includegraphics[scale=0.7]{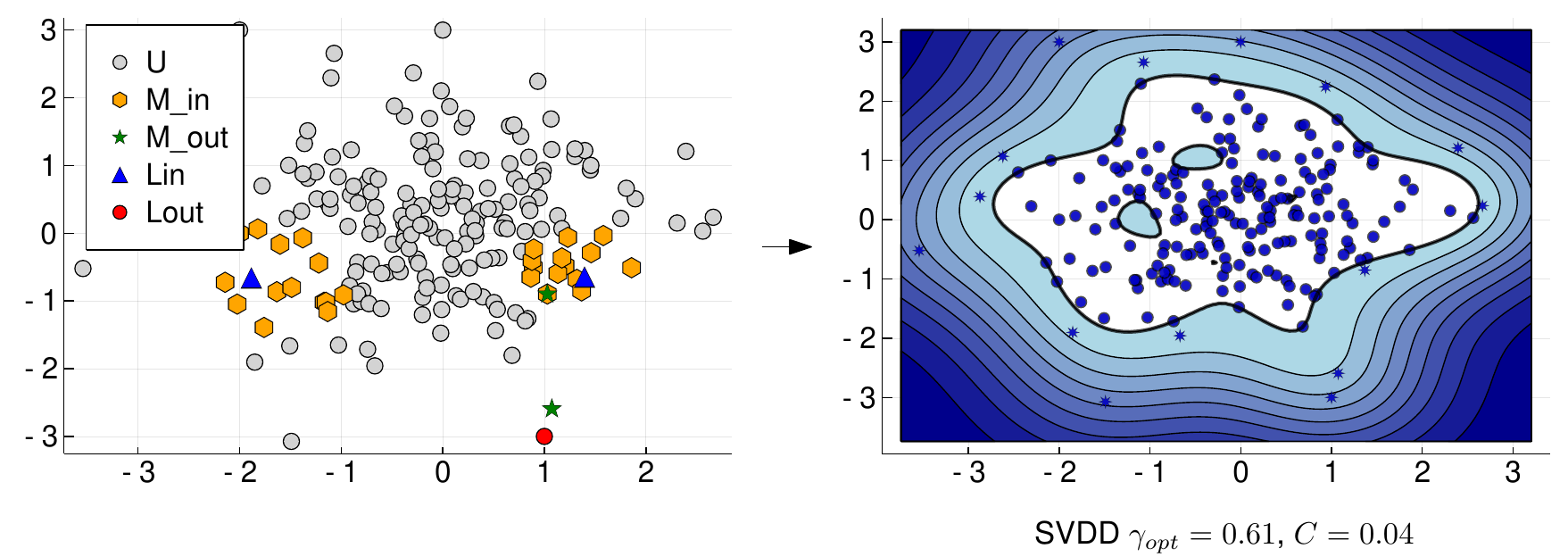}
        \subcaption{Locally optimal alignment with $k=15$.}
        \label{fig:stability-illustration-local}
    \end{subfigure}
    \caption{Comparison of global and local alignment and fitted SVDD with $\lvert\mathcal{L}_{\text{in}}\rvert = 2$ and $\lvert\mathcal{L}_{\text{out}}\rvert = 1$.}
    \label{fig:stability-illustration}
\end{figure*}

\subsubsection{Locally Optimal Alignment}

The global kernel alignment relies on all entries of the kernel matrix, see Equation~\ref{eq:alignment}.
This is problematic because, when the sample size is small and biased towards some area of the data space, $\gamma_{\text{opt}}$ may be far off the true optimum.
Figure~\ref{fig:stability-illustration} illustrates this issue on data sampled from a Gauss distribution with two labeled inliers and one labeled outlier.
In Figure~\ref{fig:stability-illustration-global}, the alignment is \enquote{global}, i.e., does not rely on neighborhood information.
It results in a large value for $\gamma_{\text{opt}}$ and causes the SVDD classifier to overfit.
In Figure~\ref{fig:stability-illustration-local}, the alignment is \enquote{local}, i.e., includes the local neighborhood of labeled observations in the alignment calculation.
The result is a small $\gamma_{\text{opt}}$ -- a good choice for the data.

We now explain how to calculate the alignment on a subset of the kernel matrix entries.
In general, an inlier should be similar to its nearest neighbors.
However, inliers may not be similar to all other inliers.
If there only are two distant observations $x_i$ and $x_j$ with $x_i, x_j \in \mathcal{L}_{\text{in}}$, a global kernel alignment would result in a large $\gamma_{\text{opt}}$, such that $k(x_i,x_j)$ is close to $1$.
In this case, $\gamma_{\text{opt}}$ overfits to the labeled observations.
To avoid this issue, we only expect inliers to be similar to their nearest neighbors that are also labeled as inliers.
Next, inliers should be dissimilar to nearest neighbors that are labeled as outliers.
Formally, this means to select the kernel matrix entries
\begin{equation}\label{eq:M-in}
    M_{\text{in}} = \{(i,j) \, | \, i \in \mathcal{L}_{\text{in}},\, j \in \mathcal{L}' \cap \text{NN}_k(i)\}
\end{equation}
With outliers, one cannot assume similarity to their nearest neighbors, since the nearest neighbor of an outlier may often be an inlier.
Thus, we assume that outliers are similar only to their symmetric nearest neighbors.
Further, outliers should be dissimilar to the nearest inliers that are not their reverse nearest neighbors.
Formally, this means to select the kernel matrix entries
\begin{equation}\label{eq:M-out}
    \begin{split}
        M_{\text{out}} = \{(i,j) \, | & \, i \in \mathcal{L}_{\text{out}}, \, 
        j \in  \left(\mathcal{L}'_{\text{out}} \cap \text{SNN}_k(i)\right)\\
        & \cup \left(\mathcal{L}'_{\text{in}} \cap \text{NN}_k(i) \setminus \text{RNN}_k(i) \right)  \} 
     \end{split}
\end{equation}
Figure~\ref{fig:stability-illustration-local} highlights $M_{\text{in}}$ and $M_{\text{out}}$.
To calculate an alignment on these subsets, we set the remaining kernel matrix entries to $0$, i.e., they do not have any impact on the alignment calculation.
\begin{equation}\label{eq:K-opt-prime-masked}
    K'_{\text{opt}}(i,j) \leftarrow 0, \quad \forall \, (i,j) \notin M_{\text{in}} \cup M_{\text{out}}
\end{equation}
\begin{equation}
    K_{\gamma}(i,j) \leftarrow 0, \quad \forall \, (i,j) \notin M_{\text{in}} \cup M_{\text{out}}
\end{equation}
We denote the alignment on this subset as
\begin{equation}\label{eq:subset-alignment}
	a_{\text{local}} := A(K_{\gamma}, K'_{\text{opt}}).
\end{equation}

\subsection{Query Strategy}\label{sec:query-strategy}

Active learning is an iterative approach to selecting observations for which users are asked to provide a class label~\cite{Settles2010-qa}.
The core idea is to improve upon some objective, usually classification accuracy, with a minimal number of iterations.
In each iteration, a \emph{query strategy} ranks observations by their informativeness, i.e., the expected benefit of knowing the class label given a specific objective.
Formally, this is a function $\tau$ that maps an observation $x \in \mathcal{U}$ to $\mathbb{R}$.
Users are asked to provide the class label for the observation with the highest informativeness, $q = \argmax_{x \in \mathcal{U}} \tau(x)$.

Active learning in combination with kernel learning has only been studied for non-parametric kernels~\cite{Hoi2008-ph}.
Conventional active learning methods are also not useful for learning hyperparameter values.
Most conventional query strategies are not applicable since they rely on already parameterized classifiers~\cite{Trittenbach2018-zp}.
Other query strategies rely only on data characteristics and select observations in the margin between classes~\cite{Ghasemi2011-tv}, i.e., they select border cases for fine-tuning an already parameterized classifier, which tend to not be representative of the underlying distribution.
Hyperparameter estimation requires observations that are informative of both classes and of the data distribution.
To our knowledge, there currently is no query strategy with the objective to estimate hyperparameter values.

In our scenario, an observation is informative if its label contributes towards finding $\gamma_{\text{opt}}$.
Intuitively, these are the observations that fit least to the current alignment, and thus lead to large changes.
The rationale is that this query strategy results is explorative at first, which leads to large changes in the alignment.
Over time, the changes become smaller, and the parameter estimation more stable.
Thus, we propose to estimate informativeness of an instance by calculating how much the alignment changes when the label for a yet unlabeled instance would become available.

\paragraph{Min-Max Alignment Query Strategy}\label{sec:MMA}

Given a current $\gamma_{\text{opt}}$, and the respective alignment $a_{\text{local}}$ both derived by Equation~\ref{eq:subset-alignment}, for each potential query $x \in \mathcal{U}$, there are two cases.
If $x$ is an inlier, the updated sets are $\mathcal{L}''_{\text{in}} = \mathcal{L}'_{\text{in}} \cup \{x\}$, otherwise $\mathcal{L}''_{\text{out}} = \mathcal{L}'_{\text{out}} \cup \{x\}$.
One must then update $M_{\text{in}}$ and $M_{\text{out}}$ respectively to calculate an updated alignment.
If $x$ is inlier, the updated alignment is $a_{\text{local}}^{\text{in}}$, otherwise it is $a_{\text{local}}^{\text{out}}$.
We define the informativeness as the minimum change in the alignment over both cases
\begin{equation}\label{eq:MMA}
	\tau_{\text{MMA}} (x) = \min \{\lvert a_{\text{local}} - a_{\text{local}}^{\text{in}} \rvert, \lvert a_{\text{local}} - a_{\text{local}}^{\text{out}} \rvert \}.
\end{equation}
So $q$ is the unlabeled observation where $\tau_{\text{MMA}}$ is maximal.
Algorithm~\ref{alg:al-kernel} is an overview of our proposed active learning method to estimate the kernel parameter.
For efficiency, we calculate $\tau_{\text{MMA}}$ on a candidate subset $\mathcal{S} \subseteq \mathcal{U}$ with sample size $\lvert \mathcal{S} \rvert$, which we select randomly in each iteration.
In our experiments, we have found a sample of size $\lvert \mathcal{S}\rvert = 100$ to work well.

\subsection{Estimating Cost Parameter C}\label{sec:estimating-c}

Active learning results in a ground truth of size $\lvert \mathcal{L} \rvert = k$ after $k$ iterations.
The sample obtained through Min-Max Alignment gives way to a grid search for $C$, as follows.
First, there is a lower bound $C_{\text{LB}}$ and an upper bound $C_{\text{UB}}$ on the feasible region of $C$.
Recall that, with decreasing $C$, more observations may fall outside of the hypersphere.
To obtain $C_{\text{LB}}$, we use binary search for the smallest $C$ where the SVDD optimization problem still has a feasible solution.
To obtain $C_{\text{UB}}$, we search for the smallest $C$ where all observations, regardless of their label, are classified as inlier.
We then use a grid search to find $C_{\text{opt}}$.
We train several classifiers in $[C_{\text{LB}}, C_{\text{UB}}]$ and compare their classification accuracy on $\mathcal{L}$ based on a suitable metric, e.g., Cohen's Kappa.
This is the \emph{quality score} that assesses the current parameter estimates.
$C_{\text{opt}}$ is the value that yields the highest score.


\begin{algorithm}[ht]
	\DontPrintSemicolon
	
	\caption{Active Learning of Kernel Parameter}
	\label{alg:al-kernel}
	
	\SetKwInOut{Data}{Data}
	\SetKwInOut{Parameter}{Parameter}
	\SetKwInOut{Output}{Output}
	
	\SetKwData{query}{q}
	
	\SetKwFunction{askOracle}{askOracle}
	\SetKwFunction{relabel}{relabel}
	\SetKwFunction{calculate}{calculate}
	\SetKwFunction{drawInitialSample}{drawInitialSample}
	
	\Data{$\ \mathcal{X} = \langle x_1, x_2, \dots, x_N \rangle$}
	\Parameter{$k$}
	\Output{$\gamma_{\text{opt}}$}
	
	\vspace{1ex}
	
	\SetKwFunction{minMaxAlignment}{$\tau$}
	\SetKwProg{minmax}{Function}{:}{}
	\minmax{\minMaxAlignment{$x$; $a$, $\mathcal{L}'_{\text{in}}$, $\mathcal{L}'_{\text{out}}$}}{
		$L''_{\text{in}} \gets L'_{\text{in}} \cup \{x\}$
		
		$L''_{\text{out}} \gets L'_{\text{out}} \cup \{x\}$
		
		\calculate $a_{\text{local}}^{\text{in}}$,  $a_{\text{local}}^{\text{out}} $ \tcp*[r]{Eq. \ref{eq:subset-alignment}, \ref{eq:MMA}}
		\Return $\min(\lvert a_{\text{local}} - a_{\text{local}}^{\text{in}}  \rvert, \lvert a_{\text{local}} - a_{\text{local}}^{\text{out}}) \rvert$ 
	}

	\vspace{1ex}
	
	$\mathcal{L_{\text{in}}}$, $\mathcal{L_{\text{out}}}$ $\leftarrow$ \drawInitialSample\; 
	
	$\mathcal{U}$ $\leftarrow$ $\mathcal{X} \; \setminus$ $\mathcal{L_{\text{in}}} \cup \mathcal{L_{\text{out}}}$\;
	
	\vspace{1ex}
	
	\While{$\neg$terminate}{
		
		$\mathcal{L}'_{\text{in}}$, $\mathcal{L}'_{\text{out}}$ $\leftarrow$ \relabel($\mathcal{L_{\text{in}}}$, $\mathcal{L_{\text{out}}}$) \tcp*[r]{Eq.~\ref{eq:relabel}}
		
		\calculate $M_{\text{in}}$, $M_{\text{out}}$ \tcp*[r]{Eq. \ref{eq:M-in}, \ref{eq:M-out}}
		
		\calculate $K'_{\text{opt}}$ \tcp*[r]{Eq. \ref{eq:K-opt-prime}, \ref{eq:K-opt-prime-masked}}
		
		\calculate $\gamma_{\text{opt}}$ \tcp*[r]{Eq. \ref{eq:gamma-opt}, \ref{eq:subset-alignment}}

		\vspace{1ex}
		\tcp*[h]{Min-Max Alignment Strategy}\;
		$a_{\text{local}} \gets A(K_{\gamma_{\text{opt}}}, K'_{\text{opt}})$
		
		$s \gets 0$
		
		\For{$x \in \mathcal{U}$}{
			$s' \gets $ \minMaxAlignment($x$; $a_{\text{local}}$, $\mathcal{L}'_{\text{in}}$, $\mathcal{L}'_{\text{out}}$)
			
			\If{$s' > s$}{
			
				$\query \gets x$
				
				$s \gets s'$
				
			}
		}
	
		\vspace{1ex}
		\tcp*[h]{Update pools}\;
		\eIf{\askOracle(\query) == \enquote{outlier}}{
			$\mathcal{L_{\text{out}}}$ $\leftarrow$ $\mathcal{L_{\text{out}}} \cup \{\query\}$\;
		}
		{
			$\mathcal{L_{\text{in}}}$ $\leftarrow$ $\mathcal{L_{\text{in}}} \cup \{\query\}$\;
		}
		$\mathcal{U}$ $\leftarrow$ $\mathcal{U} \setminus \{\query\}$
		\vspace{1ex}
	}
	\Return $\gamma_{\text{opt}}$
\end{algorithm}

%% file: sections/experiments.tex
\section{Experiments}\label{sec:experiments}

\newcommand{\ALSample}{\text{AL-Sample}\xspace}

\begin{table*}[ht]
	\centering
	\begin{tabular}{lccccccc||c}
		\toprule
		Dataset & \OurMethod &  \OurMethod-Sample &  DFN-Fix &  DFN-Sample &   QMS &  ADS-default &  ADS-ext &  Emp.\ UB  \\
		\midrule
Annthyroid       &    0.02 &     \textbf{0.03} &     0.00 &     -0.00 &  0.00 &         0.00 &     0.00 &    0.04 \\
Cardio           &    \textbf{0.25} &     0.23 &     0.00 &      0.22 &    -- &         0.00 &     0.00 &    0.24 \\
Glass            &    \textbf{0.15} &     0.09 &     0.04 &      0.03 &    -- &         0.00 &     0.00 &    0.25 \\
Heart            &    \textbf{0.13} &     0.10 &     0.00 &      0.10 &  0.00 &         0.00 &    -0.02 &    0.13 \\
Hepatitis        &    0.05 &     0.15 &     0.00 &      \textbf{0.16} &  0.05 &         0.08 &     0.05 &    0.21 \\
Ionosphere       &    \textbf{0.66} &     \textbf{0.66} &     0.00 &      0.55 &    -- &         0.59 &     0.00 &    0.78 \\
Lymph            &    0.47 &     0.41 &     0.47 &      0.39 &    -- &         \textbf{0.48} &     \textbf{0.48} &    0.51 \\
PageBlocks       &    \textbf{0.42} &     0.35 &     0.10 &      \textbf{0.42} &  0.00 &         0.00 &     0.00 &    0.52 \\
Pima             &    0.08 &     \textbf{0.16} &     0.02 &      0.07 &  0.00 &         0.00 &     0.00 &    0.14 \\
Shuttle          &    0.06 &     \textbf{0.19} &     0.13 &      0.10 &  0.00 &         0.00 &     0.00 &    0.31 \\
SpamBase         &    \textbf{0.01} &    -0.01 &     0.00 &      0.01 &  0.00 &         0.00 &     0.00 &    0.04 \\
Stamps           &    \textbf{0.18} &     0.17 &     0.08 &      \textbf{0.18} &  0.00 &         0.00 &     0.01 &    0.21 \\
WBC              &    \textbf{0.53} &     0.50 &     \textbf{0.53} &      0.46 &  0.00 &         0.00 &     0.45 &    0.59 \\
WDBC             &    \textbf{0.38} &     0.31 &     0.34 &      0.15 &  0.00 &         0.00 &    -0.01 &    0.45 \\
WPBC             &    0.01 &     \textbf{0.04} &    -0.03 &      0.01 &  0.00 &        -0.02 &    -0.05 &    0.08 \\
Wave             &    \textbf{0.05} &     0.04 &    -0.00 &      0.02 &  0.00 &         0.00 &     0.04 &    0.11 \\
		\bottomrule
	\end{tabular}
	\caption{Result on real world benchmark data; average kappa coefficient over five repetitions; best per data set in bold.}
	\label{tab:kappa-results}
\end{table*}

We evaluate our method on an established set of benchmark data for outlier detection~\cite{Campos2016-ux}.
In this benchmark, one of the classes of the original data has been declared the outlier class.
This means that some outliers can lie in dense regions, i.e., the data contains noisy labels.
We use 16 data sets of varying size and dimensionality, normalized, with varying outlier percentage between $\SI{1}{\percent}$ and $\SI{44}{\percent}$.
Large data sets are sub-sampled to $N=2000$.
Our implementations, raw results, and notebooks to reproduce our results are publicly available.\footnote{\website}

We evaluate our active learning approach against several state-of-the-art methods to estimate SVDD hyperparameter values, and compare against a random baseline and an empirical upper bound.
We repeat each experiment five times and report the average results unless stated differently.

\emph{Active Learning.}
As an initial labeled pool, we randomly draw a sample of size $\lvert \mathcal{L}_{\text{in}} \rvert = 2$ and $\lvert \mathcal{L}_{\text{out}}\rvert = 2$.
This is a relaxed version of a cold start, and not a limitation in practice.
We apply Min-Max Alignment until $\lvert \mathcal{L} \rvert = 50$.
To speed up query selection, we only calculate $\tau_{\text{MMA}}$ on a subset of size $\mathcal{S} = 100$ in each iteration, see Section~\ref{sec:MMA}.
The locality parameter for relabeling and local alignment is $k=5$; we will discuss the impact of $k$ later.
To estimate $C$, we split $[C_{\text{LB}}, C_{\text{UB}}]$ by a grid of size $20$.

\emph{Competitors.}
We use several state-of-the-art heuristics that have outperformed other competitors in experiments conducted in the respective papers.
The first heuristic is \textit{QMS}~\cite{Ghafoori2018-ug}.
We follow the recommendation in the paper to set its parameter $k = \lceil f \cdot N \rceil$, where $f$ is an a-priori estimate of the outlier ratio, which we set to $0.05$.
The second heuristic is \textit{DFN}~\cite{Xiao2014-cc}.
We use two variants: \textit{DFN-Fix} with $C=0.05$ as recommended in the paper, and \textit{DFN-Sample} where we query the label for 50 randomly selected observations and apply grid search.
The third heuristic is \textit{ADS}~\cite{Wang2018-xd}, which uses synthetic observations.
We use the grid size recommended in the paper (\textit{ADS}) and a variant with a larger grid (\textit{ADS-Ext}).

\emph{Empirical Bounds.}
As a lower baseline for the effectiveness of our query strategy, we replace Min-Max Alignment with random sampling (\textit{\OurMethod-Sample}).
Note that the other components of our approach, i.e., selecting $\gamma$ by local alignment, and $C$ by grid search remain the same as with \OurMethod.
As an empirical upper bound, we search for hyperparameter values based on the ground truth via grid search \textit{(Emp.\ UB)}.
This is an unfair comparison; it merely sets results into perspective.
Note that this is an \emph{empirical} upper bound, i.e., instances may occur where one of the competitors yields better results, e.g., for values between the grid steps.\\

\begin{figure}[t]
	\centering
	\begin{subfigure}[c]{0.23\textwidth}
		\centering
		\includegraphics[scale=0.23]{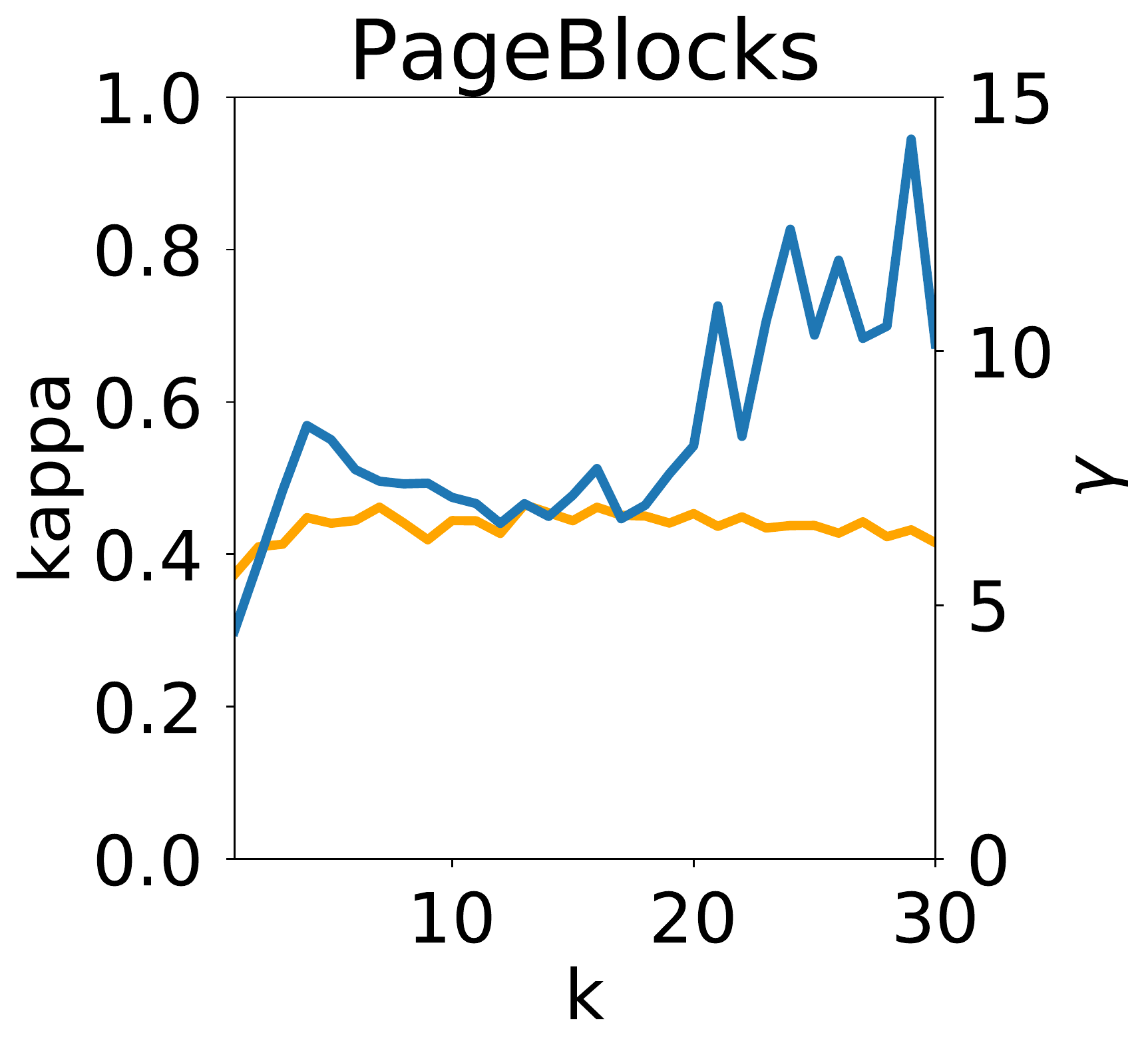}
		\label{fig:kappa1}
	\end{subfigure}
	\begin{subfigure}[c]{0.23\textwidth}
		\centering
		\includegraphics[scale=0.23]{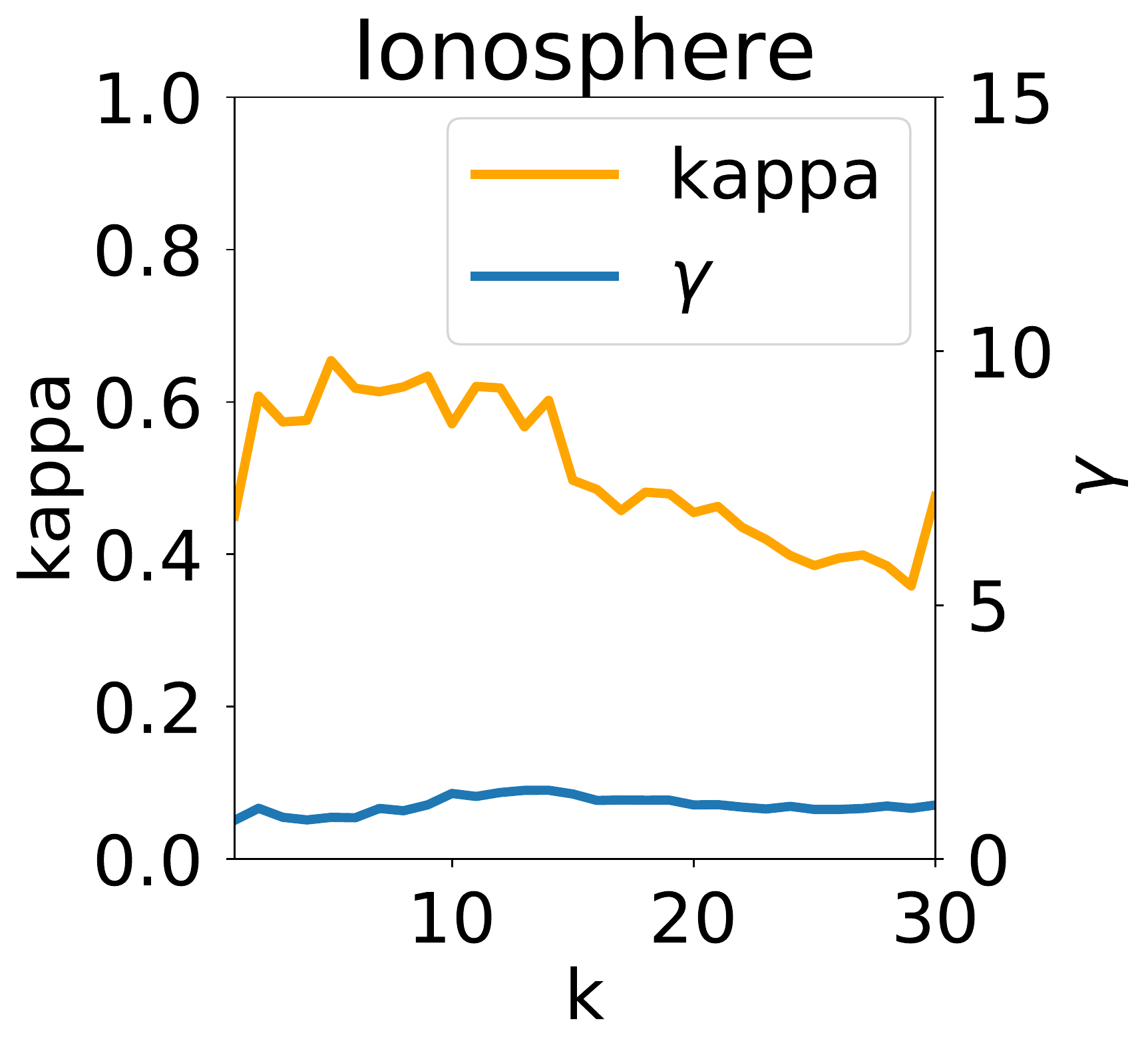}
		\label{fig:kappa2}
	\end{subfigure}
	\caption{Results with varying $k$.}
	\label{fig:kappa}
\end{figure}

One has to be careful when evaluating classification on outlier benchmark data.
This is because measuring classification quality on a holdout split assumes that the train split is representative of the data distribution; this might not hold for outliers.
We therefore suggest to evaluate on the full data set, i.e., a variant of the resubstitution error.
This a good compromise as long as there are only a few labeled observations~\cite{Trittenbach2018-zp}.
In addition, labels are only used for parameter tuning, the final classifier training is unsupervised, i.e., it does not use the obtained labels.
As evaluation metric we use Cohen's kappa, which is well suited for imbalanced data.
It returns $1$ for a perfect prediction, $0$ for random predictions, and negative values for predictions worse than random.

\OurMethod obtains very good results on the majority of the data sets, see Table~\ref{tab:kappa-results}.
In several cases, \OurMethod is even close to the empirical upper bound.
This shows that the quality score calculation on a labeled sample aligns very well with the classification quality on the full data set.
The local alignment with \OurMethod-Sample also yields good results. 
This means that our local alignment works well even on random samples.
This is in line with literature, i.e., random selection sometimes scores well against sophisticated alternatives.

\OurMethod outperforms its competitors on most data sets.
Overall, the competitors do not perform well at all, and there are only few instances where they produce useful hyperparameter values.
In many cases, the resulting classification accuracy is $0$, i.e., the competitors do not produce useful estimates in these cases.
Reasons for this might be that neither a direct estimation (DFN-Fix and QMS) nor an estimation with artificial data (ADS) works well with outlier data.
For QMS, we found that the closed formula to calculate $C$ returns values that are far from the empirical optimum.
In these cases, the classifier either classifies too many or too few observations as outliers.
Further, QMS sometimes does not return valid $\gamma$ values because of duplicates (\enquote{--}).
DFN-Sample is the closest competitor to \OurMethod.
One reason is that it uses a random sample to estimate $C$, which tends to be more effective than a closed formula.
On most data sets, however, classification results are still worse than \OurMethod-Sample, which relies on a random sample as well, but uses local kernel alignment for $\gamma$ instead of a closed formula.

We found \OurMethod to return good parameters for small values of $k$.
Figure~\ref{fig:kappa} shows the result quality and the estimated $\gamma$ value with $k$ averaged over 10 repetitions for two data sets with increasing $k$.
There, the estimated $\gamma$ as well as the result quality are stable for $k \in [5,10]$.
This means that our method is not sensitive to $k$ for small values of $k$.
In practice, we recommend to set $k=5$ since this value has worked well in our benchmark, see~Table~\ref{tab:kappa-results}.

%% file: sections/conclusions.tex
\section{Conclusions}\label{sec:conclusions}

The usefulness of SVDD largely depends on selecting good hyperparameter values.
However, existing estimation methods are purely heuristic and require a cumbersome and difficult validation of estimated values.

In this article, we propose \OurMethod, a principled approach to SVDD hyperparameter estimation based on active learning.
Its core idea is to refine kernel alignment to small sample sizes by considering only local regions of the data space.
\OurMethod provides evidence-based estimates for both SVDD hyperparameters and eliminates the need for manual validation.
\OurMethod outperforms state-of-the-art competitors in extensive experiments.
It provides estimates for both SVDD hyperparameters that result in good classification accuracy, in several cases close to the empirical upper bound.